\documentclass[11pt,a4paper]{article}
\usepackage[hyperref]{acl2019}
\usepackage{times}
\usepackage{latexsym}

\usepackage{url}
\usepackage{graphicx}
\usepackage[utf8]{inputenc}
\usepackage{amssymb}
\usepackage{fontawesome}
\usepackage{ifthen}
\usepackage{enumitem} 

\aclfinalcopy 
\def\aclpaperid{141} 

\def\placeholder#1{{\color{red}$*$#1$*$\par}}
\def\tm#1{{\color{purple}TM: #1}}
\def\noop#1{}

\let\tm\noop
\let\placeholder\noop

\title{A Test Suite and Manual Evaluation of Document-Level NMT at WMT19}

\author{Kateřina Rysová, Magdaléna Rysová, Tomáš Musil, Lucie Poláková \and Ondřej Bojar\\ 
  Charles University, Faculty of Mathematics and Physics \\
  Institute of Formal and Applied Linguistics \\
  Malostranské náměstí 25, 118 00 Prague, Czech Republic \\
  {\tt \{rysova, magdalena.rysova,  musil, polakova, bojar\}@ufal.mff.cuni.cz}}

\date{}

\begin{document}
\maketitle
\begin{abstract}
As the quality of machine translation rises and neural machine translation (NMT)
is moving from sentence to document level translations, it is becoming
increasingly difficult to evaluate the output of translation systems.

We provide a test suite for WMT19 aimed at assessing discourse phenomena of MT
systems participating in the News Translation Task.
We have manually checked the outputs and identified types of translation errors
that are relevant to document-level translation.


\tm{Doplním abstrakt o výsledky.}

\end{abstract}

\section{Introduction}

Currently, the level of machine translation systems can be very good or excellent.
For some languages, the systems are on par with humans when evaluated \emph{at
the level of individual sentences}, see \citet{microsoft:chinese:parity:2018}
for Chinese-to-English and \citet{findings:2018} for English-to-Czech
translation at WMT18. The main criterion for distinguishing MT systems' quality thus has to shift from evaluating individual sentences to larger units. Ideally, the translated text should be now evaluated as a whole.

We believe that the fundamental criterion of the quality of manual or automatic
translation is the extent to which the translation is functional in human communication.
These days, the critical basic level in this criterion has been already reached by multiple machine translation systems covering a wide range of language pairs.
While the reader of an automatically translated text may be groping at some points in the text, 
the overall quality of the translation is already so high that the main content of the text and the author's communicative intention is mostly conveyed.

Still, the reader of an MT output takes a higher effort to
understand the translated text. For example, morphological errors, shortcomings
in the word order, incorrect syntactic relations, 
failure in translating terminology, or the choice of inappropriate synonyms can
hinder the speed and accuracy of text understanding.

In this paper, we first provide a test suite for WMT19 aimed at assessing translation quality of English to Czech NMT systems regarding document-level language phenomena. As qualitative analyses of document-level errors in MT outputs are up-to-date quite rare, this paper further aims at identification, manual annotation and linguistic description of these types of errors relevant to English-Czech NMT and a comparison of performance of the submitted systems in the given areas. We compare NMT systems that translate one sentence at a time with systems that have more than one sentence on input and therefore have potential to translate document-level phenomena better.

After an overview of detected translation errors from various levels of language description, the paper zooms in on three document-level, or coherence-related, phenomena: topic-focus articulation (information structure), discourse connectives and alternative lexicalizations of connectives.\footnote{This work does not address in detail errors in coreference, pronoun and gender translation, as these phenomena have been already widely accounted for, e.g. \citet{guillou2016findings, novak2016pronoun}.} We assume that translation systems might have difficulties with these phenomena, as they are related to the previous context and go beyond (or are affected by the phenomena across) the sentence boundary. In this way, they contribute to the overall coherence of the text that should (as a whole) function as an independent unit of human communication.



\section{Data}

The evaluations in this paper are conducted on a selection of 101 documents from the parallel Prague Czech-English Dependency Treebank (PCEDT, 
\citet{pcedt20:lrec2012}), and we also used discourse annotations of the same texts in the Penn Discourse Treebank 3.0 (PDTB, for details see \citet{webber2019penn}).

\subsection{Prague Czech-English Dependency Treebank}

The Prague Czech-English Dependency Treebank is a parallel corpus consisting of English original texts and their Czech translations. The PCEDT contains 1.2 million running words in almost 50,000 sentences in each part. 

The English texts come from the Penn Treebank (Wall Street Journal Section;
\citealp{PTB}).
They were manually translated into Czech by trained linguists without any support of MT and proofread. The PCEDT is 
manually annotated on the tectogrammatical (deep-syntactic) layer in both languages. The
sentences are represented by dependency structures of content words. The nodes
in the tree structures are provided with syntactico-semantic labels as, e.g.,
predicate, actor, patiens, addressee or locative. Also, the valency frames of verbs
(argument structure) are captured, as well as elliptical structures and
anaphoric relations.   

In addition, the Czech part is automatically tagged and parsed as surface-syntactic dependency trees on the analytical layer. The English part also preserves the original phrase-structure annotation of the Penn Treebank. Also, the annotation of discourse relations, connectives and Altlexes from the Penn Discourse Treebank was extracted and added to our PCEDT dataset.


\section{NMT Systems}

We evaluated 5 NMT systems from those participating in WMT19 in English-to-Czech
translation. In particular, we selected those of the highest quality as estimated
by automatic scoring at
matrix.statmt.org.\footnote{\url{http://matrix.statmt.org/matrix/systems_list/1896}}

\texttt{CUNI-Transf-2018} is last year submission by \citet{popel2018cuni}. It is a neural machine translation model based on the Transformer architecture and trained on parallel and back-translated monolingual data. It translates one sentence at a time.

\texttt{CUNI-DocTransf-T2T} is a Transformer model  following \citet{popel2018cuni}, but trained on WMT19 document-level parallel and monolingual data. During decoding, each document was split into overlapping multi-sentence segments, where only the ``middle'' sentences in each segment are used for the final translation. \texttt{CUNI-Transf-T2T} is the same system as \texttt{CUNI-DocTransf-T2T}, just applied on separate sentences during decoding.

\texttt{CUNI-DocTransf-Marian} is document-level trained Transformer in Marian framework following \citet{popel2018cuni}, but finetuned on document-level parallel and monolingual data by translating triples of adjacent sentences at once. If possible, only the middle sentence is considered for the final translation hypothesis, otherwise a double or single sentence context is used.

\texttt{Online-B} is an anonymized online system which we know also from several previous years of WMT.

\texttt{Reference} is the Czech side of the PCEDT corpus.

\section{Annotation Design}
\label{design}

 The 101 PCEDT documents selected for translation and manual evaluation belong to the ``essay'' and ``letter'' genre labels according to the classification of PDTB given in \citet{webber2009genre}. At the same time, the selected texts have a length of 20--50 sentences. These documents were submitted as an additional test suite for Machine Translation of News shared task at the WMT 2019. 
Because we are interested in document-level translation and the effect of context on the translation, we only selected documents with cross-sentence discourse relations. 

\begin{figure*}
    \centering
    \includegraphics[width=\textwidth]{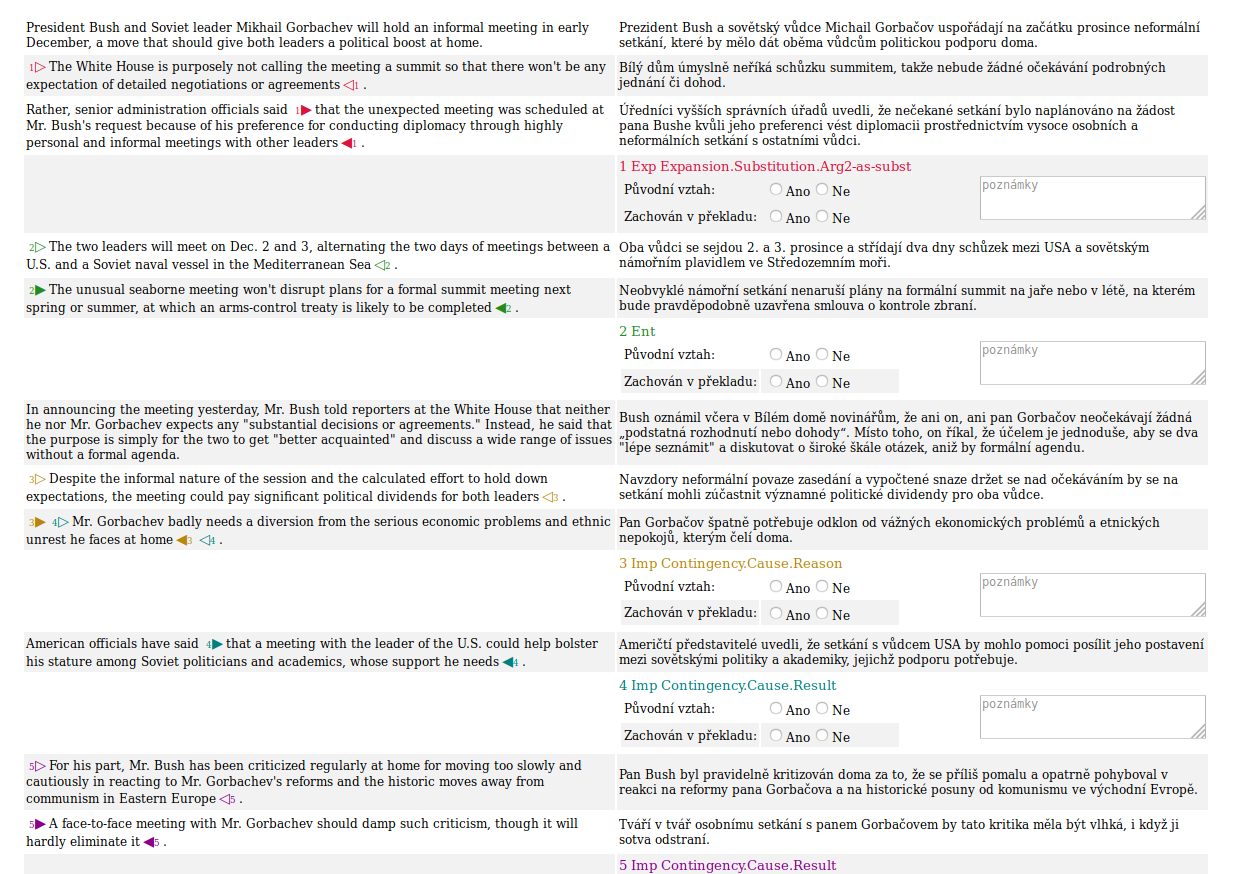}
    \caption{Screenshot of the annotation interface.}
    \label{fig:interface}
\end{figure*}

We have created a simple annotation interface (see Figure~\ref{fig:interface}), which allows the annotator to mark the items that were translated correctly.

Specifically, several types of cross-sentence discourse relations are considered on the
source side (reusing the annotations available in the Penn Discourse Treebank 3.0).

The target side was validated by trained linguists. For each of the
observed connectives / AltLex, the annotators indicated whether: 

\noindent(1) the given expression/phrase in the source fulfills the function of a connective -- according to the annotator, or the function of AltLex -- according to the original English annotation displayed. If yes, then whether its Czech translation is (2):

\begin{itemize}[noitemsep]
\item adequate and correctly placed,\footnote{for Altlexes: and preserves the original discourse meaning}
\item adequate but incorrectly placed,
\item omitted and it does not harm the output
\item omitted and it harms the output
\item not adequate.
\end{itemize}


The questionnaire for word order annotation is analogous, compare the description of tables with results below in Section~\ref{results}. 
The original translation into Czech from PCEDT could serve as a
reference translation but similarly to \citet{findings:2018}, we opted for a
bilingual evaluation, showing the annotators always the source and the candidate
translation. The benefit is that the human translation can be evaluated using
the same criteria as the MT system outputs. 

There were 6 annotators, all of them students of linguistics. Each annotator evaluated 8 documents in the first round. For each document, they evaluated the output of one MT system (without knowing which MT system produced the output). To measure the inter-annotator agreement, we organized a second round of evaluation, where each annotator was given documents and systems combination that was in the first round evaluated by another annotator. Details on the IAA are given in Section~\ref{results}.

\section{Linguistic Analysis of Translations Errors across Language Levels}

We carried out a complex linguistic analysis of a sample of the translated texts and  revealed that even the best translations contained cca 15–20 linguistic issues (per text of 35 sentences). This means that although the content reliability and
linguistic level of (the best) MT systems is very high, they still do
not reach communication skills of humans. This fact may be challenging for their
authors, as there are still possibilities for improvement. However, a systematic
improvement of MT systems is rather difficult due to non-systematic nature of language
errors found in the analysis -- e.g. if there appeared an untypical word order in a sentence, it does not mean that word order errors are also present in the rest of the translated text. It turned out, on the contrary, that the errors / problematic issues appear individually, as singularities.

In the following part, we discuss the problematic places in a sample of translated texts. We tried to select the best or (at least) good MT systems to demonstrate that even in such an advanced translation, there are still issues requiring improvement. 

\subsection{Morphology}
\label{morf}
We were able to detect errors from various levels of language description. 
Some problematic issues concerned even such basic phenomena as e.g. the use of a verbal mood or other morphological issues (\textit{It's as if France decided to give only French history questions to students in a European history class, and when everybody aces the test, they say their kids are good in European history -- Je to, jako by se Francie rozhodla dávat studentům evropských hodin dějepisu jen otázky z francouzštiny, \textbf{a když všichni v testu excelují, říkají}, že jejich děti jsou v evropských dějinách dobré}; the Czech translation is not consistent in maintaining potentiality: the intended content should be translated into Czech as: \textit{jako \textbf{kdyby} se Francie..., a \textbf{až by} všichni v textu excelovali, řekli \textbf{by}...}) with the obligatory conditional morpheme \textit{by}, also as a part of the conjunction \textit{kdyby}, used in past (unreal) conditions.

\subsection{Lexicon}

Other 
issues concerned the choice of vocabulary. The individual translations included e.g. inappropriate repetition of the same word (ie. the MT systems produced a non-natural output by not attempting to use a synonym, cf. \textit{in test-coaching workbooks and worksheets –- v pracovních sešitech a pracovních sešitech} “in test-coaching workbooks and in test-coaching workbooks”). In some of them, there also appeared incorrect literal translations of terms (cf. \textit{a joint venture of McGraw-Hill Inc. and Macmillan's  \textbf{parentt}, Britain's Maxwell Communication Corp -- společným podnikem McGraw-Hill Inc. a Macmillanovým  \textbf{rodičem}, britskou společností Maxwell Communication Corp}).  

Another lexical issue was the use of an inaccurate synonym in a given context (cf. \textit{but he doesn't deny that some items are similar -- ale nepopírá, že některé předměty jsou podobné}; the word \textit{předměty} may be a synonym to the original \textit{items} but not in this context, the Czech word here means rather tangible \textit{objects}).

Generally, the MT systems succeed in translating basic words or phrases but sometimes they fail in translating terms or technical words and in  lexical variety (often resulting in word repetition and failure to use an appropriate synonym). 

\subsection{Syntax}

The translations also exhibit signs of incorrect syntactic relations, e.g. excessive genitive accumulation, which is untypical for Czech (cf. \textit{About 20,000 sets of Learning Materials teachers' binders have also been sold in the past four years. -- Asi 20 000 souborů (Noun in Gen) učebních materiálů (NP in Gen) učitelských pořadačů (NP in Gen) bylo také prodáno v posledních čtyřech letech.}). Another typical syntactic error appears in translation of syntactically potentially homonymous phrases, as in the example above in \ref{morf} the phrase \textit{European history class}, translated wrongly as \textit{evropských hodin dějepisu} (European classes of history).

Also, a large problematic area was revealed in word order configurations. Some translations contained the word order adopted from English, where it is untypical or even incorrect in Czech. This issue is related to sentence information structure or topic-focus articulation, as the word order is connected with contextual boundness (cf. \emph{\ldots{}says "well over 10 million" of its Scoring High test-preparation books have been sold since their introduction 10 years ago -- uvádí, že ``více než 10 milionů'' jeho testovacích knih Scoring High se prodalo od jejich zavedení před 10 lety}; the expression  \emph{``více než 10 milionů''} is the focus of the sentence and therefore it should be placed in the final position in Czech). Similar issue (concerning topic-focus articulation) may be observed in the sentence \textit{Scoring High and Learning Materials are the best-selling preparation tests. -- Scoring High and Learning Materials jsou nejprodávanější přípravné testy}. Again, the expression \textit{Scoring High and Learning Materials} should be (as focus proper of the sentence) placed in the final sentence position in Czech. 

\subsection{Semantics}

Semantic issues (to a certain extent) are already partly included in the incorrect translations of terms as discussed above. Other are related especially to factual inaccuracy, e.g. the expression \textit{French history questions} was incorrectly translated as \textit{otázky z francouzštiny} “questions from French”. 

In some cases, even a whole part of the original text was completely omitted in the translation -- the meaning of the sentence was thus negatively affected (\textit{\ldots{}and Harcourt Brace Jovanovich Inc.'s Metropolitan Achievement Test and Stanford Achievement Test -- \ldots{}a Harcourt Brace Jovanovich}).

\subsection{Discourse}

Further issues in translations also appeared on higher levels of language description, crossing the sentence boundary and mostly affecting text understanding as a whole. These discourse-related phenomena include especially coreference and discourse (semantico-pragmatic) relations, largely expressed by discourse connectives or their paraphrases (AltLexes). A detailed analysis of discourse-related translation errors is given below in Section~\ref{disc_things}.


\section{Linguistic Analysis of Selected Document-Level Errors}

\subsection{Selected coherence phenomena}
\label{disc_things}

A comprehensive linguistic analysis of a sample of translated texts showed that even the best translations are not completely error-free (the best ones contained about 15--20 errors per text). These errors were further analyzed -- they appear across individual levels of language description. Unfortunately, the main common feature of the errors seems to be the fact that they are not systematic. The key to a good distinction of translation quality is thus their complex linguistic analysis. 
For the annotation, we have chosen three document-level types of the errors discovered in the output analysis, namely those concerning \textbf{topic-focus articulation, discourse connectives} and the meanings they convey and \textbf{alternative lexicalizations of connectives} (AltLexes).  
The annotators then assessed them on a larger sample of translated data from all the systems and the reference translation. The finding are analyzed linguistically in the rest of this Section and quantitatively below in Section~\ref{results}.

\subsubsection{Topic-focus articulation and word order}
First, we observed the phenomenon of topic-focus articulation (we follow this phenomenon as presented within the Functional Generative Description, see \citet{sgall} or \citet{sgall2}). In our experiment, we took advantage of the fact that English and Czech have a different word order system in combination with topic-focus articulation and contextual boundness.\footnote{For definitions of terms related to topic-focus articulation and contextual boundness see \citet{hajicova2013}. } While English has a fixed word order, strongly influenced by grammar, Czech has a free word order mainly influenced by the contextual boundness of individual sentence constituents. It is thus necessary to harmonize the word order in a Czech sentence always with respect to the previous (con)text.

In the annotation of the translated texts, we focused on the word order of the subject. While the subject is typically at the beginning of the sentence in English, it can occupy various positions in Czech, depending on whether it is contextually bound or not. We were wondering how individual MT-systems reflect this word order issue.

We automatically selected English original sentences from the PCEDT that contained a noun used with an indefinite article in the subject position and its Czech counterparts in evaluated translated texts. It is assumed that this subject is contextually non-bound (not deductible from the previous context, it is “new” information) and is thus expected elsewhere than at the beginning of the sentence, most likely to follow the predicate in Czech. Moreover, this subject (or the constituent corresponding to it in Czech) could be also so-called \textit{focus proper} standing at the very end of the Czech sentence in written texts.

For Czech translations, it was necessary to check whether the Czech equivalent of the English subject was retained as a contextually non-bound sentence constituent and whether it was appropriately located in the Czech sentence, see the following example.\\

English text: \textit{What is the best-selling preparation test? A NEW LANGUAGE TEST is the best-selling preparation test.}\\

Expected Czech translation: \textit{Co je nejprodávanějším přípravným testem? Nejprodávanějším přípravným testem je NEW LANGUAGE TEST.}

\subsubsection{Discourse connectives and their sentence positions}

The second phenomenon assessed in the annotation were discourse connectives. Discourse connectives are rather short function words (e.g. \textit{but, therefore, nevertheless, because,} or \textit{and}) that connect two text units while expressing a discourse (semantico-pragmatic) relation between them, thus ensuring text to a large extent text coherence and cohesion.
Here, the problematic issues included the use of a wrong Czech equivalent -- both from the semantic and grammatical point of view (e.g. the positions of connectives in a sentence etc.). An example of a wrong connective translation is as follows. 
\textit{\textbf{Since} chalk first touched slate, schoolchildren have wanted to know: What's on the test?} --
\textit{\textbf{*Protože} se křída poprvé dotkla břidlice, žáci chtěli vědět: Co je na testu?}
    
The English connective \textit{since} is homonymous and its meaning may be causal or temporal. In the example, it was translated as causal (by the Czech connective \textit{protože -- because}) in a temporal context (the correct Czech translation here would be \textit{od okamžiku, kdy} (from the moment when...). 
Such an incorrect translation of a discourse connective demonstrates nicely the potential huge impact on overall comprehensibility.

From the word order perspective, even these cohesive devices have their typical positions in a clause -- according to their part-of-speech classification. Coordinating conjunctions typically stand between two discourse units (\textit{I play the flute \textbf{and} I dance.} / \textit{Hraju na flétnu \textbf{a} tančím.}) both in English and Czech. Subordinating conjunctions typically occur at the beginning of the discourse unit to which they belong syntactically (\textit{Because it rains, I'm not going out. I won't go out because it rains.} / \textit{\textbf{Protože} prší, nepůjdu ven. Nepůjdu ven, \textbf{protože} prší.}). Connectives of adverbial origin have looser positions in some cases;\footnote{For more information see \citet{rysova2018}.} they can occur e.g. in the first and second position in the sentence (\textit{For me it is easier to not lose a game than to win it, \textbf{thus} I produce better results in stronger tournaments. Both umpires claimed that they were unsighted, and were \textbf{thus} forced to give Somny the benefit of the doubt.} / \textit{Pro mě je snazší neztratit hru, než ji vyhrát, \textbf{proto} dosahuji lepších výsledků v silnějších turnajích. Oba rozhodčí tvrdili, že neviděli, byli \textbf{proto} nuceni dát Somnymu výhodu pochybovat.}).

In some word-order positions of discourse connectives, English and Czech differ. In other words, a Czech translation should not copy the connective ordering from an English original. In English, some discourse connectives can occur e.g. at the very end of the sentence (cf. \textit{too, as well, instead, nevertheless} etc.), which is not typical for Czech. 

To better compare the quality of the individual translations, we observed especially the translation equivalents of multi-word connectives like \textit{as long as} or \textit{as much as} that could be problematic due to their idiomatic character.

\subsubsection{Alternative lexicalizations of discourse connectives (AltLexes)}

In addition to discourse connectives, discourse relations can also be expressed by their alternatives called AltLexes, see \citet{prasad2010}. Alternative lexicalizations of connectives are often multi-word phrases such as \textit{for this reason}. Since these cohesive structures often have an idiomatic character and they generally do not achieve such degree of grammaticalization as connectives, their forms in languages may vary to a large extent. 

For example, the AltLex \textit{for this reason} is not translated into Czech literary as \textit{pro tento důvod} ‘lit. for this reason’, but as \textit{z tohoto důvodu} ‘lit. from this reason’. Other examples of English AltLexes are \textit{that’s all, that’s largely due to, attributed that to, it will cause} etc. A list of AltLexes in English is given in \citet{prasad2007}, multi-word connective expressions in Czech are described and presented in \citet{rysovakniha}. Due to their high lexical variety and lower degree of grammaticalization, AltLexes were selected for the annotation as potentially interesting expressions for translation.

\section{Results}
\label{results}

In this section, we present the results of the evaluation.

\def\stars#1{%
\ifthenelse{\equal{#1}{4.5}}{\faStar\faStar\faStar\faStar\faStarHalfO}{%
\ifthenelse{\equal{#1}{4.0}}{\faStar\faStar\faStar\faStar\faStarO}{%
\ifthenelse{\equal{#1}{3.5}}{\faStar\faStar\faStar\faStarHalfO\faStarO}{%
\ifthenelse{\equal{#1}{3.0}}{\faStar\faStar\faStar\faStarO\faStarO}{%
\ifthenelse{\equal{#1}{2.5}}{\faStar\faStar\faStarHalfO\faStarO\faStarO}{%
\ifthenelse{\equal{#1}{2.0}}{\faStar\faStar\faStarO\faStarO\faStarO}{%
\ifthenelse{\equal{#1}{1.5}}{\faStar\faStarHalfO\faStarO\faStarO\faStarO}{%
\ifthenelse{\equal{#1}{1.0}}{\faStar\faStarO\faStarO\faStarO\faStarO}{%
\ifthenelse{\equal{#1}{0.5}}{\faStarHalfO\faStarO\faStarO\faStarO\faStarO}{%
\ifthenelse{\equal{#1}{0.0}}{\faStarO\faStarO\faStarO\faStarO\faStarO}{}%
}}}}}}}}}%
}

\def\tresaltlex{
\begin{tabular}{lccc}
 & adequate & missing & wrong \\
CUNI-Transf-2018       & \stars{4.5} & \stars{0.0} & \stars{0.5} \\
CUNI-DocTransf-Marian  & \stars{3.5} & \stars{0.5} & \stars{1.0} \\
online-B               & \stars{4.0} & \stars{0.0} & \stars{1.0} \\
CUNI-DocTransf-T2T     & \stars{4.0} & \stars{0.0} & \stars{1.0} \\
CUNI-Transf-2019       & \stars{4.0} & \stars{0.0} & \stars{1.0} \\
reference              & \stars{3.5} & \stars{0.0} & \stars{1.5} \\
\end{tabular}}

\def\treskonekt{
\begin{tabular}{lcccc}
 & a & ax & m & n \\
CUNI-Transf-2018      & \stars{4.5} & \stars{0.5} & \stars{0.0} & \stars{0.0} \\
CUNI-DocTransf-Marian & \stars{4.0} & \stars{1.0} & \stars{0.0} & \stars{0.0} \\
online-B              & \stars{4.5} & \stars{0.5} & \stars{0.0} & \stars{0.0} \\
CUNI-DocTransf-T2T    & \stars{4.5} & \stars{0.5} & \stars{0.0} & \stars{0.0} \\
CUNI-Transf-2019      & \stars{4.5} & \stars{0.5} & \stars{0.0} & \stars{0.0} \\
reference             & \stars{4.0} & \stars{0.0} & \stars{0.0} & \stars{0.5} \\
\end{tabular}}

\def\treskonektP{
\begin{tabular}{lccc}
 & adequate & wrong place & inadequate \\
CUNI-Transf-2018      & \stars{4.5} & \stars{0.5} & \stars{0.0} \\
CUNI-DocTransf-Marian & \stars{4.0} & \stars{1.0} & \stars{0.0} \\
online-B              & \stars{4.5} & \stars{0.5} & \stars{0.0} \\
CUNI-DocTransf-T2T    & \stars{4.5} & \stars{0.5} & \stars{0.0} \\
CUNI-Transf-2019      & \stars{4.5} & \stars{0.5} & \stars{0.0} \\
reference             & \stars{4.0} & \stars{0.0} & \stars{0.5} \\
\end{tabular}}

\def\tresccontext{
\begin{tabular}{lcc}
 & yes & no \\
CUNI-Transf-2018 & 11 & 1 \\
CUNI-DocTransf-Marian & 17 & 3 \\
online-B & 6 & 1 \\
CUNI-DocTransf-T2T & 17 & 1 \\
CUNI-Transf-2019 & 6 & 1 \\
reference & 19 & 4 \\
\end{tabular}}

\def\tresrema{
\begin{tabular}{lcc}
 & yes & no \\
CUNI-Transf-2018 & 2 & 0 \\
CUNI-DocTransf-Marian & 5 & 3 \\
online-B & 0 & 1 \\
CUNI-DocTransf-T2T & 1 & 2 \\
CUNI-Transf-2019 & 0 & 1 \\
reference & 1 & 6 \\
\end{tabular}}

\def\tresrazeni{
\begin{tabular}{lcc}
 & yes & no \\
CUNI-Transf-2018 & 14 & 0 \\
CUNI-DocTransf-Marian & 14 & 5 \\
online-B & 3 & 1 \\
CUNI-DocTransf-T2T & 13 & 3 \\
CUNI-Transf-2019 & 6 & 0 \\
reference & 19 & 3 \\
\end{tabular}}

\def\tresccrr{
\begin{tabular}{lcc|cc|cc}
 & \multicolumn{2}{c}{context} & \multicolumn{2}{c}{rhema} & \multicolumn{2}{c}{placement} \\
 & yes & no & yes & no & yes & no \\
 \hline
CUNI-Transf-2018      & 11 & 1 & 2 & 0 & 14 & 0 \\
CUNI-DocTransf-Marian & 17 & 3 & 5 & 3 & 14 & 5 \\
online-B              & 6  & 1 & 0 & 1 & 3 & 1 \\
CUNI-DocTransf-T2T    & 17 & 1 & 1 & 2 & 13 & 3 \\
CUNI-Transf-2019      & 6  & 1 & 0 & 1 & 6 & 0 \\
reference             & 19 & 4 & 1 & 6 & 19 & 3 \\
\end{tabular}}

\subsection{Inter-annotator agreement}

The inter-annotator agreement was measured pairwise, it ranges from 66~\% to 93~\% with an average of 80~\%. The agreement was on average 69\,\%  for AltLexes, 87\,\% for connectives and 79\,\% for questions concerning word order.

\subsection{AltLexes}

The annotation interface for alternative lexicalizations contained identical questions to those for connective assessment (described above in Section~\ref{design}), with the exception of their (in)correct placement, as this question is irrelevant for such non-grammaticalized phrases. There were 23 queries in average for each of the evaluated translations. The results for adequacy of AltLex translations in each system output AND the reference are summed up in Table~\ref{tab:altlex}.
\placeholder{LP: já ty hvězdičky asi nechápu. Když je každá 10 procent, neměly by na řádku sčítat do 10? Tak snad to interpretuju dobře, přečtěte si to po mně radši} 
\tm{Já jsem si uvědomil, že taky nevím, jak si to přesně s těmi hvězdičkami Ondra představoval, ale až když jsem je začal dávat do těch tabulek\ldots{} Momentálně se to nesčítá na 100, protože je to zaokrouhlené na nejbližších 10\,\% a to je půlhvězdička, ne hvězdička (jak jsem omylem původně psal výše).}

A source AltLex was assessed as an appropriate connecting device in accordance with the original discourse annotation in 130 cases (Yes), and inappropriate in 42 cases (No). The proportion of negative answers is surprisingly high, but a closer look on the data reveals that the annotators, quite in unity (but in contrast to the PDTB notion of AltLex), resist treating \textbf{verbs} as a specific form of connecting devices. This mostly concerns causative verbs like \textit{to explain, to strengthen} or \textit{to blame}. They might be in fact right, these verbs are mostly translated well and their role in discourse coherence is a rather supplementary one. 
Apart from this issue, Table~\ref{tab:altlex} demonstrates that once an AltLex is approved as a connecting device, it is in vast majority of cases translated correctly (rarely incorrectly), the original discourse meaning is preserved and it is not omitted in the translation. This applies quite equally across all systems, with a small decrease for CUNI-DocTransf-Marian system and the reference (!). A potential explanation is the typically looser human translation (and possibly the context-aware Marian system).


\def\hvezdaexpl{Each $\bigstar$ represents 20\,\% and the results are rounded to the nearest half-star.}

\begin{table*}[ht]
    \centering
    \tresaltlex{}
    \caption{Results for AltLex annotations.  \hvezdaexpl{}}
    \label{tab:altlex}
\end{table*}

\subsection{Connectives}

As for connectives, there were 52 queries in average for each of the evaluated translations. The results for adequacy of connective translations in each system output and the reference are summed up in Table~\ref{tab:konekt}.
A source connective candidate was assessed as an factual connecting device 
in 303 cases (Yes), and not a connective in 30 cases (No). This proportion seems to be correct, the non-connective readings of some expressions are relevant, e.g. several times for \textit{as much as} in the function (and position) of a quantifier. 
Once a connective candidate is approved as an actual connective, it translated always correctly (compare column ``n" in Table~\ref{tab:konekt}), but it is possibly incorrectly placed in the translation (column ``ax"). The result figures indicate that there are no significant differences across the systems in translating the traced connectives.





\begin{table*}[ht]
    \centering
    \treskonekt{}
    \caption{Results for connectives annotations. The columns are: (a) adequate and correctly placed, (ax) adequate but incorrectly placed, (m) omitted and it does not harm the output, and
(n) not adequate. \hvezdaexpl{}}
    \label{tab:konekt}
\end{table*}


\subsection{Word order}

\tm{V téhle podsekci nechávám absolutní čísla místo "hvězdiček", protože mi vzhledem k velikostem a poměrům připadá zkreslující to hvězdičkovat
\\
TODO: tabulky tady se asi sloučí do jedné}

The word order evaluation focused the translation of contextually non-bound subjects (representing a new information in the sentence). The annotators first determined, which of the automatically preselected sentences from the English source indeed contain a contextually non-bound subject (85 Yes, 10 No). If yes, they traced whether the subject in the Czech translation also contextually non-bound. The results of manual annotation demonstrate that MT systems in general preserve the contextual non-boundness of the subjects. The figures are comparable across the systems, only the Marian system and the reference achieved a slightly worse scores:

\tresccontext{}




In a second task, we observed whether the subject in the English original the focus proper of the given sentence. Again, the annotators first filtered out relevant sentences (10 Yes, 36 No). Then they looked at whether the subject in the Czech translation is also the focus proper of the sentence. Similarly as in the previous task, the Marian system's performance is worse, and the performance of CUNI-DocTransf-T2T drops. However, the results here are less significant, as there were only few occurrences of the annotated tokens:



\tresrema{}

Next, we followed the systems' ability to place the Czech equivalents of the original English subjects correctly into the Czech output sentence. Here, a correct placement according to the Czech word order rules was mostly achieved by all systems. There was not enough data collected for the online-B system, but the rest is comparable, with both context-aware systems performing slightly worse than others:

\tresrazeni{}



\section{Conclusion}

In this paper, we have described a test suite of parallel English-Czech texts provided for WMT19 with the aim to assess discourse phenomena in output of MT systems participating in the News Translation Task.
We have carried out an extensive manual annotation of the MT outputs and identified types of translation errors relevant to document-level translation. We also compared the systems' performance with respect to the observed phenomena. 

In general, the recent NMT systems have achieved such a high level of translation quality that it has become difficult to evaluate their output in a systematic fashion. Most of the
errors in the translation cannot be found by a simple comparison with the reference translation, a bilingual evaluation is needed. Moreover, for the observed phenomena, the systems performed with only a minor differences among each other and they reached the quality of the reference. In fact, the reference translation was in some aspects evaluated as worse, which is likely caused by the greater literal adherence of the automatic translations to the original and it does not mean that the reference is incorrect. Contrary to our assumptions, the two context-aware systems did not outperform the others in translating the followed document-level phenomena. This can be attributed to the fact that the systems perform good enough on this task already, and also partly because the evaluation can change a lot using just a slightly different annotation setting, e.g. if we traced also other (ambiguous) connective expressions or anaphoric items. The actual errors are difficult to predict from scratch and they occur randomly.
More specifically, while the translations of AltLexes and discourse connectives showed quite satisfactory (at least of those observed here), the most errors (equally across systems) were detected in the area of word order and contextual (non-)boundness of the subjects. The systems prefer to keep the original word also in the translations, not really accounting for the impact of information structure. 

\section*{Acknowledgement}

We acknowledge support from the Czech Science Foundation project no. GA17-06123S
(Anaphoricity in Connectives: Lexical Description and Bilingual Corpus Analysis)
and the EU project H2020-ICT-2018-2-825460 (ELITR). This study has utilized language resources distributed by the LINDAT/CLARIN project of the Ministry of Education, Youth and Sports of the Czech Republic (project LM2015071).
This research was partially supported by SVV
project number 260 453.

\bibliography{main}
\bibliographystyle{acl_natbib}

\end{document}